\documentclass{clv2}

\pdfoutput=1

\usepackage{amsmath}
\usepackage{multirow}
\usepackage{url}
\usepackage{enumitem}
\usepackage{bm}

\runningtitle{POSLDA}
\runningauthor{Darling \& Song}

\begin{document}

\title{Probabilistic Topic and Syntax Modeling with Part-of-Speech LDA}

\author{William Darling\thanks{6 chemin de Maupertuis, 38240 Meylan, France. E-mail: \tt{william.darling@xrce.xerox.com}.}}
\affil{Xerox Research Centre Europe}

\author{Fei Song\thanks{50 Stone Road East, Guelph, Ontario, N1G 2W1, Canada. E-mail: \tt{fsong@uoguelph.ca}.}}
\affil{University of Guelph}

\maketitle

\begin{abstract}
This article presents a probabilistic generative model for text based on semantic topics and syntactic classes called Part-of-Speech LDA (POSLDA). POSLDA simultaneously uncovers short-range syntactic patterns (syntax) and long-range semantic patterns (topics) that exist in document collections. This results in word distributions that are specific to both topics (sports, education, ...) and parts-of-speech (nouns, verbs, ...). For example, multinomial distributions over words are uncovered that can be understood as ``nouns about weather'' or ``verbs about law''. We describe the model and an approximate inference algorithm and then demonstrate the quality of the learned topics both qualitatively and quantitatively. Then, we discuss an NLP application where the output of POSLDA can lead to strong improvements in quality: unsupervised part-of-speech tagging. We describe algorithms for this task that make use of POSLDA-learned distributions that result in improved performance beyond the state of the art.
\end{abstract}

\section{Introduction}

Two highly related phenomena are resulting in a renaissance for the study of computational linguistics. The first is the increasing level of access to textual data sources over the Internet such as classic works of literature (The Gutenberg Project), structured explanatory knowledge of the world (Wikipedia), people's every thought (Twitter), governments' every desire (laws and legal decisions), and ongoing triumphs, defeats, changes, and breaking information (online news). The second, concomitant with the first, is the growth of interest in, and the power of, machine learning algorithms which can exploit the vast amounts of data that are being made available and help make sense of them. 

Like language itself, machine learning techniques can be described and contrasted with each other along a number of different axes of understanding and dichotomies. One of the most important of these dichotomies is the division between \textit{supervised} and \textit{unsupervised} learning approaches \cite{hlda}. While supervised approaches are concerned with generalizing a function that predicts an output $y$ given some input $x$ learned from examining labeled example pairs $(x_i,y_i)$, unsupervised approaches involve uncovering hidden patterns and associations that exist in data \cite{elements}. Clustering algorithms, for example, are unsupervised machine learning techniques that attempt to group data together because they are similar in some way. A logical definition of similarity -- especially with respect to linguistics -- is, informally, that two texts are similar because they discuss the same topics.

Another canonical example of unsupervised learning is dimensionality reduction. In reducing the dimensionality of a dataset, an algorithm seeks to find a simpler or more condensed representation of the data while preserving (or bringing forth) some kind of meaning. An apt dimensionally-reduced representation of texts -- collections of words -- is the topics that they represent. In the standard bag-of-words representation used for many natural language processing tasks, the cardinality of the dimensional representation is the number of distinct words that can be used in the texts, that is, the vocabulary. This number is typically on the order of several thousand, while a text's content, in a broad sense, can also be described by the topics that it addresses out of a finite number of generic topics on the order of hundreds or less.

It is therefore no surprise that probabilistic topic models, which are generative models of text (based on unsupervised machine learning algorithms), are continually growing in interest. They provide a means to uncover the hidden thematic structure that underlies large document collections and therefore allow us to explore, summarize, and understand what a collection is about with ease and efficiency. Latent Dirichlet Allocation (LDA) \cite{lda}, the original topic model, describes the generative story that lays the foundation for this kind of model. It posits that a document can be created through a random process of drawing words from a mixed-membership model. At a high level, a document is created by first selecting the topics that it will address, and then randomly generating words from distributions associated with those topics. Numerous more complex models have been presented that build on this basic idea and several introductory papers exist that cover the area in detail \cite{blei2012}.

More specifically, the LDA generative process works as follows. For each document $d$, a document-specific topic portion $\theta_d$ is drawn from a Dirichlet distribution. $\theta_d$ is a discrete distribution over $K$ topics and corresponds to the weight that each topic will have in the document. Then, for each word $w_i$, a topic index $z_i$ is drawn from $\theta_d$. To generate the word, a token is drawn from a topic-specific word distribution $\phi^{(z_i)}$. There are $K$ topic-specific word distributions, each of which corresponds to a distribution over words specific to the given topic. To generate a document, however, one would require the word distributions specific to each topic, and for each document, one would also need the topic portion. Neither of these are readily available from the input, but they can be learned from a document collection by reversing the generative process through posterior inference.

In unsupervised learning, we cannot simply write a general algorithm to find what we hope to be interesting patterns. The ``no free lunch'' theorem tells us essentially that no interesting patterns can be uncovered if we do not assume that certain kinds of patterns must exist \cite{nofreelunch}. It turns out that assuming that documents reflect latent topics and that words from the same topics will co-occur is a good assumption and thus interesting results can be learned through reversing the assumed model such as the words that are most important to each topic, and a dimensionally-reduced representation of each document in topic space. However, the correct number -- or the type of assumptions -- is important. If we make too many, the data may not fit those assumptions and the output will be nonsense. Conversely, if we make too few, interesting patterns may be missed.

The standard document representation for topic modeling is the bag-of-words \cite{salton}. Each document is represented by the number of times that each word in a fixed vocabulary appears. Because word order is ignored, a great deal of meaning is lost, but the representation is efficient and has proved to be successful. However, word order -- ignored in canonical topic models -- is clearly of importance to language because though some sense may be extracted from a text whose words have been scrambled, the full meaning can only come about through a parsing of the words based on their order and relation to each other. It has also been shown that different parts of the brain are used to understand semantics and syntax \cite{stm}. Traditional topic models miss this kind of syntactical information because they are never exposed to word-order patterns in the first place. While the bag-of-words approach is efficient, further advances in NLP will require algorithms to be able to have the full understanding that humans are afforded.

In this article we introduce a new probabilistic generative model called Part-of-Speech LDA (POSLDA) which considers both the semantic topic that a word is associated with (if any) and its syntactic purpose in a sentence. This approach allows a more structured view of language creation and as a result, the posterior distributions that are learned are more specific and meaningful than in other topic models. It also allows NLP applications to make better predictions and deductions because each piece of information -- syntactic and semantic -- provides evidence that can help disambiguate the purpose and meaning of a word.

The article is organized as follows. In section 2, we discuss previous work that has focused on bringing syntactic information into probabilistic topic models. In section3, we explain our model, POSLDA, in detail. We then present the results of three sets of experiments in section 4: first, we demonstrate qualitatively the interpretability of the uncovered posterior distributions with example syntax-specific topics on a number of diverse datasets; second, we report quantitative results on the model's ability to generalize on unseen texts and to uncover high quality topics; and third, we show how the model's ability to disambiguate word use through the joint influences of semantics and syntax can lead to better results in unsupervised part-of-speech (POS) tagging than a Bayesian Hidden Markov Model (HMM). Finally, in section 5, we conclude with thoughts on future work.

\section{Topics and Syntax}

The constraints that are imposed by language on phrase structure and word order are called syntax \cite{snlp_manning}. The syntactic meaning of a word helps to explain its functional purpose in a sentence, whereas the semantic meaning is related to its lexical-thematic purpose. The former is based on short-range dependencies at the sentence level, while the latter realizes long-range dependencies at the document level. LDA and other topic models uncover patterns by exploiting the long-range dependencies of words co-occurring. Here, we want to add the short-range dependency structures to the model and we do so by focusing on modeling the \textit{functional} purpose of a word in a sentence. We are therefore interested in the part-of-speech category that a word -- in its given context -- belongs to. These include nouns, verbs, adjectives, adverbs, prepositions, conjunctions, etc. The canonical tool for unsupervised word syntax modeling is the hidden Markov model (HMM) \cite{hmm_tut} and it is therefore a natural place to begin in adding syntax information to a semantic topic model.

The first work in combining syntactic notions of language with probabilistic topic models is based on embedding an LDA-like model in a single state of an HMM \cite{hmmlda}. This model -- dubbed HMMLDA -- represents an asymmetric composite model where all generated words follow short-range syntactic dependencies, but only ``semantic'' words that are generated from a single state obey long-range dependencies. Words that carry long-range dependencies will be generated given the document-specific topic distribution, and other words will be generated independent of the current theme. This framework is different from the traditional use of the HMM in syntax modeling as each state will not correspond to a discrete part-of-speech. The content class in particular -- which is designated as the sole class that can generate ``semantic'' words -- will need to subsume nouns, verbs, adjectives, and other words that are topic-dependent. We will return to this simplifying assumption when we present our POSLDA model.

More formally, the HMMLDA model is defined by two sets of sequential latent variables $(z_n)_{n=1}^{N}$, which represent the latent topics for each word, and $(c_n)_{n=1}^{N}$, which represent the latent syntactic classes for these words. One state in the model, $s_0 \in S$, is designated as the semantic class where the LDA-like topic model is embedded. Each topic $k$ is associated with a discrete topic-word distribution $\phi^{(k)}$ and each class $s \ne s_0$ is associated with a syntax word distribution $\phi^{(s)}$. Like LDA, each document $d$ can be described by a distribution over topics $\theta^{(d)}$. However, unlike LDA, each word $w_i$ only depends on its topic $z_i$ if its class $c_i = s_0$. A word's class is modeled with the embedded HMM and transitions between classes $s_{i}$ and $s_{i+1}$ are encoded in a transition matrix $\bm{\pi}$. The generative process by which a document is created under the HMMLDA model is as follows.

\begin{enumerate}
\item{Draw $\theta^{(d)} \sim$ Dirichlet($\alpha$)}
\item{For each word $w_i$ in document $d$}
\begin{enumerate}
\item{Draw topic $z_i \sim \theta^{(d)}$}
\item{Draw class $c_i \sim \pi^{(c_{i-1})}$}
\item{If $c_i=s_0$:}
\begin{enumerate}
\item{Draw $w_i \sim \phi^{(z_i)}$}
\end{enumerate}
\item{Else:}
\begin{enumerate}
\item{Draw $w_i \sim \phi^{(c_i)}$}
\end{enumerate}
\end{enumerate}
\end{enumerate}

Like LDA, exact posterior inference is intractable for the HMMLDA model \cite{hmmlda}. Griffiths, et al.\ therefore turn to Gibbs sampling. The HMM in the HMMLDA is a Bayesian HMM meaning that the transition rows $\pi_r$ and the emission probabilities $\phi^{(c)}$ are multinomial random variables with Dirichlet priors. The same framework for collapsed Gibbs sampling can therefore be used as with the original LDA.

One of the most interesting qualities of the HMMLDA model is that stop-words and other ``syntax-only'' words are pulled to the non-semantic classes so that the learned topics are interpretable and noise-free without any need for pre-processing or \textit{a priori} stop-word removal. This defines one of the key motivations in the development of POSLDA. While the model seems to learn distributions that are noticeably useful, it is still far from perfect. It almost exclusively finds nouns as content words when verbs are often equally as important in semantic topics. This could be due to the use of the HMM that learns to discern nouns from other parts-of-speech and simply sees other semantically-important types of words as outliers that are pushed to the syntactic classes.

Another recent approach at combining syntax and semantics into a coherent probabilistic generative model is the Syntactic Topic Model (STM) \cite{stm}. Unlike the HMMLDA model, where a word is deemed to either come from a corpus-wide syntax class \textit{or} a semantic-based topic, the STM discovers topics that are both syntactically and semantically coherent. This is more directly in line with our goals for the POSLDA model. As for the motivation in combining these notions of word information, Boyd-Graber and Blei provide an edifying example:

\begin{verbatim}
Next weekend, you could be relaxing in ________.
\end{verbatim}
There are two distinct text-modeling based approaches one could use to reason about what word might be used to fill in the blank. With a topic model such as LDA, where this document is about travel, high probability words might include ``sailing'', ``Rome'', or ``flight''. With a syntax model, it might determine that the missing word should be a noun, and thus high probability words could include ``bed'', ``church'', or ``school''. However, as is explained in \cite{stm}, the best candidate to fill in the blank is an intersection of these two types of reasoning. The word ``sailing'' matches the topic related aspect of the sentence (travel), but does not fit syntactically (verb). The opposite is the case for the noun ``school''. However, when both syntactic and semantic notions of language are taken into account, a word such as ``Rome'', which has high probability both in the travel topic and in the noun syntax class, will be selected with high probability.

The STM is a non-parametric model where the number of topics is not set \textit{a priori} but is determined, through posterior inference, by the data. While the standard LDA model draws the document topic portions for a document from a $K$-dimensional Dirichlet distribution with a fixed value of $K$ topics, here the transition distributions $\pi$ and document topic portions $\theta_d$ are drawn from a Dirichlet Process (DP), where a vector $\beta$ of infinite length is a global weight that is drawn from a stick-breaking distribution and is used as a base measure for the DP. This approach frees users of the model from having to determine themselves how many topics a corpus might contain. Our POSLDA model can also easily become a nonparametric Bayesian model by incorporating a hierarchical Diriclet Process (HDP) prior \cite{hdp}.\footnote{The approach to do so specifically for POSLDA is outlined in \S 3.4 of \cite{darlingphd}.} In this formulation, the number of topics can be learned from the data in the sense that through inference we can learn the optimal number of topics $K$ that will maximize the likelihood of the data.

\begin{table}
\begin{footnotesize}
\begin{center}
\caption{Example Topics from the STM}
\label{stmt}
\begin{tabular}{|c|c|c|c|c|}
\hline
hates & bucks & runs & professor & stock\\
dreads & surges & falls & phd & share\\
mourns & climbs & walks & candidate & mutual\\
fears & falls & sits & grad & fund\\
despairs & runs & climbs & student & on\\
\hline
\end{tabular}
\end{center}
\end{footnotesize}
\end{table}

The generative process for the STM is as follows:
\begin{enumerate}
\item{Draw global weights $\beta \sim \text{GEM}(\alpha)$}
\item{For each topic index $k = \{1,...\}$:}
\begin{enumerate}
\item{Draw topic $\tau_k \sim \text{Dir}(\sigma \rho_u)$}
\item{Draw transition distribution $\pi_k \sim \text{DP}(\alpha_T, \beta)$}
\end{enumerate}
\item{For each document $d = \{1,...,M\}$:}
\begin{enumerate}
\item{Draw document weights $\theta_d \sim \text{DP}(\alpha_D, \beta)$}
\item{For each sentence root node with index $(d,r) \in \text{SENTENCE-ROOTS}_d$:}
\begin{enumerate}
\item{Draw topic assignment $z_{d,r} \propto \theta_d \pi_{start}$}
\item{Draw root word $w_{d,r} \sim \tau_{z_r}$}
\end{enumerate}
\item{For each additional child with index $(d,c)$ and parent with index $(d,p)$:}
\begin{enumerate}
\item{Draw topic assignment $z_{d,c} \propto \theta_d \pi_{z_{d,p}}$}
\item{Draw word $w_{d,c} \sim \tau_{z_{d,n}}$}
\end{enumerate}
\end{enumerate}
\end{enumerate}
While this generative process is clearly much more complex than the corresponding one for simpler models such as LDA and HMMLDA, it is also more powerful. The key steps are 3(b)(i) and 3(c)(i) where the topic assignment for a word is chosen to be a convolution of the long-range semantic topic portion $\theta_d$ and the short-range syntactic probability $\pi_{z_{d,p}}$. Some example semantically and syntactically coherent topics learned from the STM are shown in Table \ref{stmt}. Note that in each case, the high probability words are both syntactically equivalent (noun, verb, etc.) and semantically related in a topic modeling sense.

While the posterior distributions learned by the STM appear to fall in line with our goals of a syntactically-cognizant generative topic model, it suffers from certain limitations that we would like to address. First, the generative story depends on a form of meta-sentence structure that exists before the words have been generated. That is, texts are generated based on the probabilities in sentence-specific dependency parse trees \cite{stm}. Therefore, to perform inference with the STM -- and recover the posterior distributions like those in Table \ref{stmt} -- the data must be separately pre-processed into sentence dependency parses. This means that the model is not learning the syntax patterns itself, but is using information that must be supplied by a separate algorithm before inference is performed. Conversely, we are interested in a fully generative model that is consistent across syntax and semantics where inferring the short-range sentence-wide dependencies forms part of the model.

In the next section we will present our model, POSLDA. It is a strong generalization of HMMLDA in that it follows the idea that we can combine an HMM with an LDA-like topic model, but it takes the idea further so that topic-dependent words can also be influenced by different syntax classes and thus learn part-of-speech specific posterior distributions.

\section{POSLDA}

While LDA is in some sense a simple extension of probabilistic latent semantic analysis \cite{plsa}, it can be seen as the first fully generative topic model by virtue of the Dirichlet prior that is placed on the document-topic portions which in effect free the model from specific training data. Since its inception, LDA has been extended in numerous ways and particularly by infusing the model with additional factors. Word distributions can become more specific if we consider that generated words are dependent on not only the current topic, but also other latent aspects such as the sentiment of the writing and the writer's personality or ideological perspective \cite{PaulGirju10,ahmedxing}. This allows one to uncover such word distributions as ``positive/negative words about films'' or ``words about weather from the perspective of Americans/Swedes/Australians''. In fact, this approach is so powerful that it has been generalized into techniques that can easily add specific factors to topic models through the use of strong prior information \cite{paul2012}. However, to include word syntax, we need a different approach because this factor does not come out of the \textit{types} of words that are used, but their \textit{order}.

Part-of-Speech LDA (POSLDA) is an extension and generalization of LDA and HMMLDA that is designed to understand the long- and short-range dependencies between words, and as a tool for more complex NLP tasks that require both semantic and syntactic information to attain optimal performance. In an HMM, words are considered independent of their wider context within a document, but depend on the classes of the words that appear before them. Therefore, the word order in a syntax model is important and the bag-of-words representation used in canonical topic models is no longer appropriate. Because both types of word information are important, and modeling each separately entails certain restrictions, we seek to bridge these restrictions with a unified model of language, POSLDA.

Under POSLDA, each word token is now associated with two latent variables: a topic $z$ and a syntactic class $c$. We posit that the topics are generated through the LDA process, while the classes are generated through a Bayesian HMM. The observed word tokens are then generally dependent on both the topic and the class: rather than a single discrete distribution for a particular topic $z$ or a particular class $c$, there are distributions for each topic-class pair $(z,c)$ from which we assume words are sampled. However, there also exists a set of ``syntactic-only'' words that do not depend on the thematic context of a document \cite{hmmlda}. These words -- such as determiners, prepositions, and conjunctions -- are often called ``function'' words and should be modeled as ``universal'' syntax classes that are not affected by -- and are not assigned -- a latent topic.

We therefore are interested in a generative model of text where all generated words depend on the function that they perform in a sentence, and a subset of these words also depend on the current semantic topic. For the HMM-like portion of the model, we denote the set of classes $\mathcal{C} = \mathcal{C}_{\textsc{sem}} \cup \mathcal{C}_{\textsc{syn}}$, which includes the set of semantic classes $\mathcal{C}_{\textsc{sem}}$ and the set of syntactic (function word) classes $\mathcal{C}_{\textsc{syn}}$. If a word is generated from a function word class, it does not depend on the topic. This allows our model to accommodate functional words that appear independently of the topical content of a document.

\begin{figure}
\begin{center}
\includegraphics[width=2.5in]{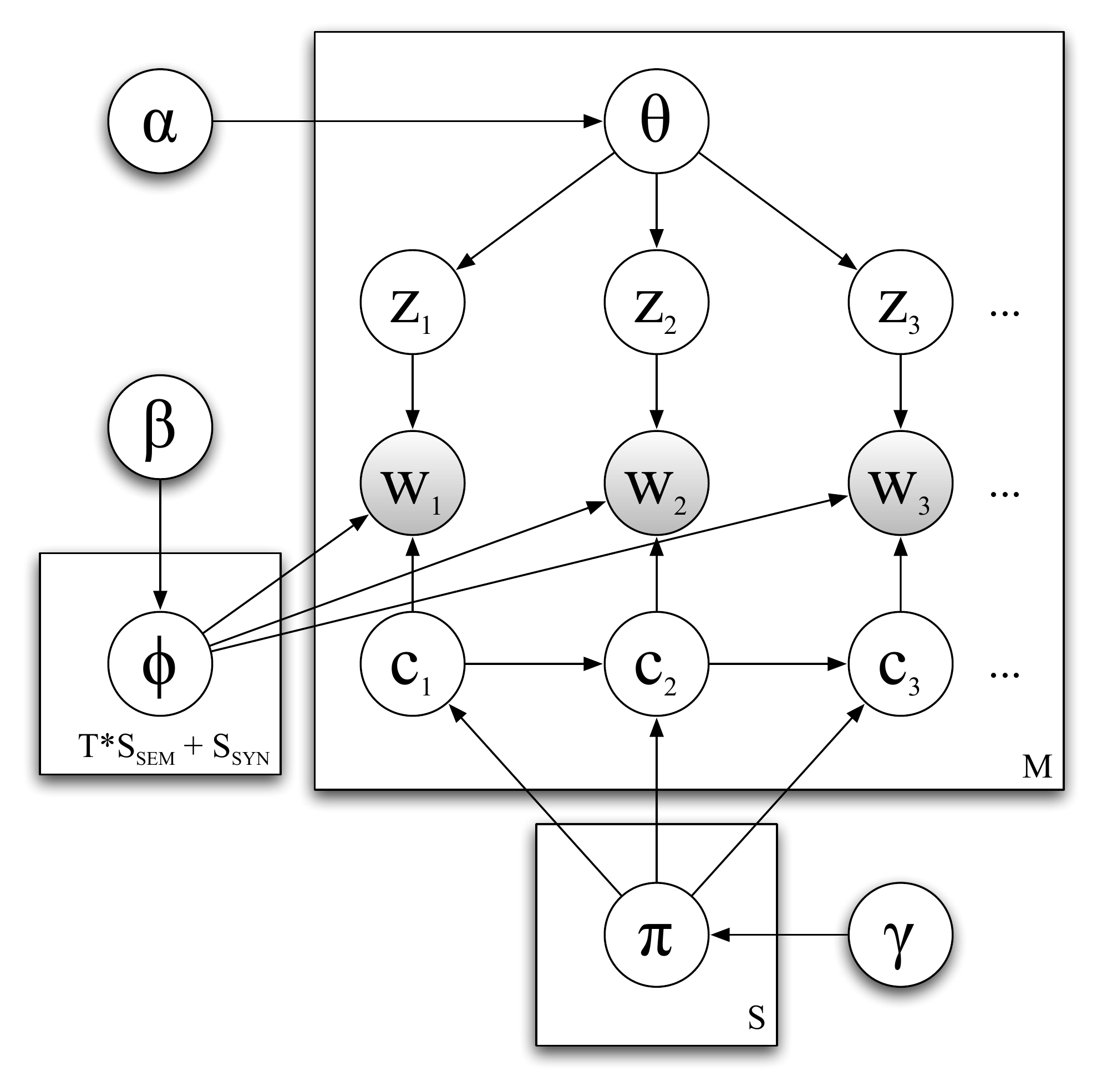}
\includegraphics[width=2.5in]{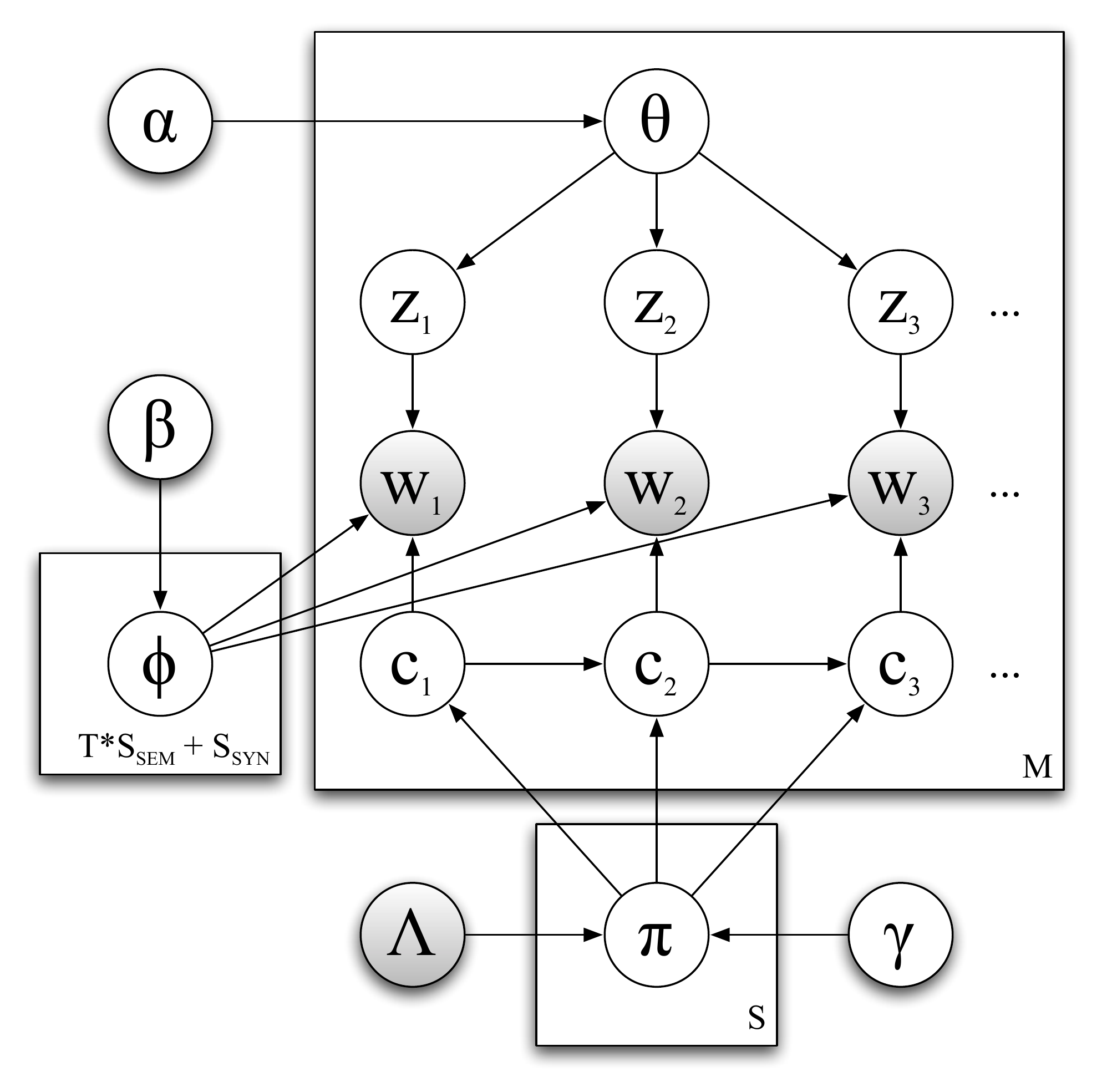}
\end{center}
\caption{\label{fig:poslda_gm} Graphical model depiction of POSLDA (left) and Labeled POSLDA (right) for unsupervised POS tagging.}
\end{figure}

We use a similar notation to LDA, where $\theta_d$ is a document-topic portion and $\phi_\eta$ is a word distribution. Additionally, we denote the HMM transition matrix $\bm{\pi}$, which we assume has rows that are drawn from a Dirichlet distribution with hyperparameter $\gamma$. Denote $S=|\mathcal{C}|$ and $T=|\mathcal{Z}|$, the numbers of classes and topics, respectively. There are $S_{\textsc{syn}}$ word distributions $\phi^{(\textsc{syn})}$ for function word classes and $T \times S_{\textsc{sem}}$ word distributions $\phi^{(\textsc{sem})}$ for semantic classes. A graphical model depiction of POSLDA is shown in Figure \ref{fig:poslda_gm}. This figure also denotes a slight variation to the model called Labeled POSLDA which is analogous to Labeled LDA for the original LDA \cite{labeledlda}. Here, an observed dictionary denoted by $\Lambda$ restricts the classes that certain words can take on. We will make use of this model for dictionary-based unsupervised part-of-speech tagging.

The generative process for a corpus of documents $\mathcal{D}$ under the POSLDA model is described as follows:
\begin{enumerate}
\item{For each row $\pi_r \in \bm{\pi}$:}
\begin{enumerate}
\item{Draw $\pi_r \sim \text{Dirichlet}(\gamma)$}
\end{enumerate}
\item{For each word distribution $\phi_{\eta} \in \bm{\phi}$:}
\begin{enumerate}
\item{Draw $\phi_{\eta} \sim \text{Dirichlet}(\beta)$}
\end{enumerate}
\item{For each document $d \in \mathcal{D}$:}
\begin{enumerate}
\item{Draw $\theta_d \sim \text{Dirichlet}(\alpha)$}
\item{For each word token $w_i \in d$:}
\begin{enumerate}
\item{Draw $c_i \sim \pi_{c_{i-1},\dots,c_{i-n}}$}
\item{If $c_i \notin \mathcal{C}_{\textsc{sem}}$:}
\begin{enumerate}
\item{Draw $w_i \sim \phi^{({\textsc{syn}})}_{c_i}$}
\end{enumerate}
\item{Else:}
\begin{enumerate}
\item{Draw $z_i \sim \theta_d$}
\item{Draw $w_i \sim \phi^{({\textsc{sem}})}_{c_i,z_i}$}
\end{enumerate}
\end{enumerate}
\end{enumerate}
\end{enumerate}
In traditional topic models, it is generally the case that common function words will overwhelm the word distributions, leading to suboptimal results and learned word distributions that are difficult to interpret. This problem is often skirted by either data pre-processing (e.g. removing stop words from a domain-dependent list) \cite{lda}, backing off to ``background'' word models \cite{general_specific,PaulGirju10}, or by performing term re-weighting \cite{WilsonChew10}. In the case of POSLDA, these common words are naturally explained by the corresponding function word classes and are pushed to these distributions rather than the topic-specific distributions during learning.

\subsection{Relations to Other Models}

The idea of having discrete word distributions for the cross product of topics and classes is related to multi-faceted topic models where word tokens are associated with multiple latent variables \cite{PaulGirju10,ahmedxing,paul2012}. Under such models, words can be explained by a latent topic as well as a second (or $n$th) underlying variable such as the perspective or dialect of the author, and words may depend on both (or multiple) factors. In our case, the second variable is the part-of-speech -- or functional purpose -- of the token.

POSLDA is also similar to a recent model called Nested HMM-LDA (nHMMLDA) \cite{nhmmlda}. The model described is very similar to POSLDA but contains certain limitations. Principally, rather than allowing each word to be generated from any of $K$ topics, all words from a sentence must share the same topic. This is a strong assumption since it will not allow a sentence to discuss more than a single topic.

POSLDA is constructed in a generalized manner and contains many existing models as special cases. For example, POSLDA reduces to a Bayesian HMM when the number of topics $K = 1$, the original LDA model when the number of classes $S = 1$, or the HMMLDA model when the number of semantic classes $S_{\textsc{sem}}=1$. One of the key benefits of these reductions is that one can easily experiment with any of these models using a single POSLDA implementation by simply altering the necessary parameters. POSLDA can also easily reduce to the nHMMLDA model by forcing all words in a sentence to share the same topic.

\subsection{Approximate Inference}

The principal computational problem in probabilistic topic/syntax models is posterior inference \cite{blei2009topic}. As it is based on LDA and the Bayesian HMM, exact inference in the POSLDA model is also intractable. Therefore, following many others, we make use of the MCMC-based approximate inference technique collapsed Gibbs sampling \cite{griffiths04finding,heinrich2004parameter}. Here, the multinomial parameters are first integrated out and we directly sample the indexing latent variables $c_i$ and $z_i$. In POSLDA, if the class is designated as \textit{syntactic}, then it only depends on the class. We therefore introduce the counts $n_{w_i}^{(c_i,z_i)}$ which correspond to the number of times that word $w_i$ is assigned to class $c_i$ \emph{and} topic $z_i$. Our sampling equation is then as follows:

\begin{equation}
\label{eq:sampling_eq}
p(c_i, z_i | \bm{c_{-i}}, \bm{z_{-i}}, \bm{w}) \propto
\begin{cases}
\rho_{c_i} \times \frac{n_{z_i}^{(d)} + \alpha_{z_i}}{n_{.}^{(d)}+ \alpha_{.}} \frac{n_{w_i}^{(c_i,z_i)} + \beta}{n_{.}^{(c_i,z_i)} + W\beta} & c_i \in C_{\textsc{sem}}\\
\rho_{c_i} \times \frac{n_{w_i}^{(c_i)} + \beta}{n_{.}^{(c_i)} + W\beta} & c_i \in C_{\textsc{syn}}
\end{cases}
\end{equation}
where
\begin{equation}
\rho_{c_i} = \frac {n_{(c_{i-2},c_{i-1},c_{i})} + \gamma_{c_i}} {n_{(c_{i-2},c_{i-1})} + \gamma_{.}} \cdot \frac {n_{(c_{i-1},c_{i},c_{i+1})} + \gamma_{c_i}} {n_{(c_{i-1},c_{i})} + \gamma_{.}} \cdot \frac {n_{(c_{i},c_{i+1},c_{i+2})} + \gamma_{c_i}} {n_{(c_{i},c_{i+1})} + \gamma_{.}}
\end{equation}

Note that we sample the pair $(c_i,z_i)$ jointly as a block, which requires computing a sampling distribution over $S_{\textsc{syn}} + T \times S_{\textsc{sem}}$.  It would also be valid to sample $c_i$ and $z_i$ separately, which would require only $S+T$ computations, in which case, the sampling procedure would be somewhat different. Despite the lower number of computations per iteration, however, the sampler is likely to converge faster with our blocked approach because the two variables are tightly coupled. The intuition is that a non-block-based sampler could have difficulty escaping local optima because we are interested in the most probable \emph{pair}; a highly probable class $c$ sampled on its own, for example, could prevent the sampler from choosing a more likely pair $(c', z)$.

\begin{figure}
\begin{center}
\includegraphics[width=3.5in]{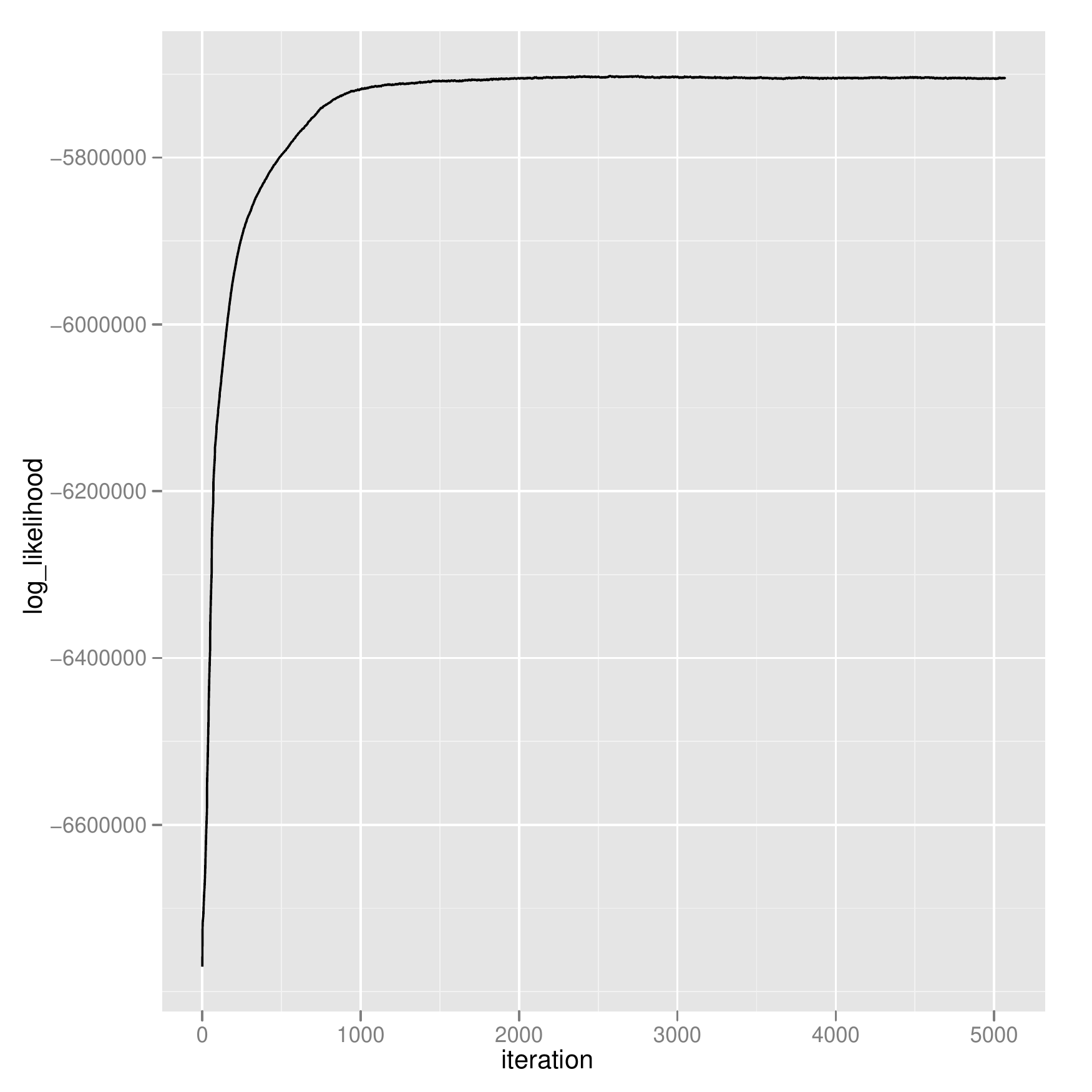}
\end{center}
\caption{\label{fig:gibbs_converge} Gibbs iterations vs.\ log-likelihood while learning a POSLDA model from the TREC AP dataset with $K=50$, $S=10$, $S_{\textsc{sem}}=5$.}
\end{figure}

One problem with MCMC-based methods is that assessing convergence can be difficult. We do not address specific approaches to get around this issue, but we note that a stabilizing likelihood can be used to infer convergence. Generally this happens after several hundred iterations (depending on the size of the dataset), and for the datasets used in this article, the likelihood will converge at about 2,000 iterations. Finally, we are interested in deriving point estimates for the topics $\phi^{(c,z)}$ from the sampled statistics. $\phi$ is a 3-dimensional array where $\phi_{w_i}^{(c,z)} = p(w_i | c, z)$. Therefore, following \eqref{eq:sampling_eq}, we get
\begin{equation}
p(w_i | c, z) = \frac{n_{w_i}^{(c,z)} + \beta}{n_{.}^{(c,z)} + W\beta}
\end{equation}

\section{Experiments and Results}

In this section we present a set of experiments on the POSLDA model to demonstrate its capabilities as a topic and syntax model of language. We demonstrate both qualitatively and quantitatively the model's ability to capture the semantic and syntactic axes of information prevalent in a corpus. We begin qualitatively with topic interpretability and then present quantitative results on the ability of POSLDA as a predictive language model. Following this, we show its ability as a model for performing unsupervised POS tagging.

\subsection{Topic Interpretability}

Judging the interpretability of a set of topics is highly subjective. Chang, et al.\ look at ``word intrusion'' where a user determines an \emph{intruding} word from a set of words that does not thematically fit with the other words, and ``topic intrusion'' where a user determines whether the learned document-topic portion $\theta_d$ appropriately describes the semantic theme of the document \cite{tealeaves}. Here, we are mostly interested in subjectively demonstrating the low incidence of ``word intrusion'' both in terms of semantics (theme) and syntax (part-of-speech). We subjectively demonstrate that our model learns semantic and syntactic word distributions that are likely robust towards problems of word intrusion.\footnote{This is in line with our approach of viewing topic models as \emph{tools} for performing other tasks. We are most interested in objective quantitative results learned from applying our models to NLP tasks such as POS tagging which we demonstrate later in this section.}

\begin{table}
\begin{center}
\begin{footnotesize}
\begin{tabular}{|c|c|c|c|c|c|c|c|c|}
\hline
\multicolumn{3}{|c|}{\textbf{\textit{``law''}}} & \multicolumn{3}{c|}{\textbf{\textit{``finance''}}} & \multicolumn{3}{c|}{\textbf{\textit{``health''}}} \\
\textit{adj} & \textit{verb} & \textit{noun} & \textit{adj} & \textit{verb} & \textit{noun} & \textit{adj} & \textit{verb} & \textit{noun} \\
\hline
federal & filed & attorney & stock & rose & exchange & health & died & study\\
court & ruled & judge & wall & averaged & stock & medical & suffered & research\\
supreme & agreed & district & bond & issued & securities & aids & received & hospital\\
legal & contends & calif & million & fell & dow & drug & underwent & virus\\
civil & claims & county & american & gained & york & blood & found & report\\
appeals & contended & board & financial & dropped & inc & heart & carried & disease\\
tax & refused & loan & composite & rated & totaled & research & suffers & university\\
illegal & sued & san & common & traded & drexel & immune & leaves & doctor\\
government & won & court & business & stocks & commission & hospital & kills & person\\ 
financial & wrote & justice & dow & closed & lambert & cancer & took & patient\\
\hline
\end{tabular}
\end{footnotesize}
\caption{\label{examples} Example topics learned from the TREC AP dataset with POSLDA.}
\end{center}
\end{table}

To demonstrate the effectiveness of POSLDA's semantic-syntactic pattern recognition ability, we fit a number of models to different datasets. We demonstrate results both on more ``traditional'' news corpora such as TREC AP and the WSJ, and on more esoteric datasets such as collections of tweets from the microblogging website \textit{Twitter} and collections of legal decisions from the Supreme Court of Canada.\footnote{\url{http://scc.lexum.org/en/index.html}.} We begin with a standard demonstration of topic interpretability on news data from the Associated Press.

\subsubsection{Traditional News Data}

Table \ref{examples} shows three topics -- manually labeled as ``law'', ``finance'', and ``health'' -- learned from a 2,250 document subset of the TREC AP corpus \cite{trec}. We set the number of topics $K = 30$, the number of classes $S = 17$, and the number of semantic classes $S_{\textsc{SEM}} = 7$.\footnote{We choose this number based on intuition. We imagine that of the 17 often-delineated parts-of-speech \cite{bayesianhmm}, 6 or 7 will generally be theme-specific. These include adjectives, verbs, gerund or present participle verbs, adverbs, nouns, and past participle verbs. It can be helpful to include an ``extra'' semantic class to see if the model can find patterns of semantics-syntax that we might miss.} We show the top ten words from three POS-specific topics labeled manually as \emph{adjective}, \emph{verb}, and \emph{noun}. The interpretability of the topics and the cohesiveness of the terms with high probabilities is clear. All three topics assign high probabilities to words that one would expect to have high importance. More importantly, however, the POS-specific topics also clearly reflect their syntactic roles. Each of the verbs is assuredly (even without the proper context) a verb, and the same thing for the nouns. The adjectives seem to fit as well; though many of the words could be considered nouns depending on the context, it is clear how given the topic each of the words can very well act as an adjective. A final point worth mentioning is that, unlike LDA, we do not perform stop-word removal. Instead, the POSLDA model has pushed stop-words to their own \emph{syntactic} classes (rather than semantic) freeing us from having to perform pre- or post-processing steps to ensure interpretable topics. The top words in four of these topic-independent syntactic classes are shown in Table \ref{examples2} with manually-labeled class names.

\begin{table}
\begin{center}
\begin{footnotesize}
\begin{tabular}{|c|c|c|c|}
\hline
\sc{auxiliary} & \sc{conjunction} & \sc{determiner} & \sc{relative} \\
\hline
is & and & the & that\\
was & but & a & which\\
be & or & an & who\\
are & \& & this & when\\
has & so & some & what\\
have & both & such & how\\
will & times & any & where\\
would & nor & many & whose\\
says & plus & those & why\\
were & yet & these & whom\\
\hline
\end{tabular}
\end{footnotesize}
\end{center}
\caption{\label{examples2} Example topic-independent syntax class distributions ($\mathcal{C}_{\textsc{syn}}$) learned from the AP dataset with POSLDA.}
\end{table}

Next we look at the ACL\_DCI release of the WSJ treebank dataset which contains approximately 3 million words over 6,058 documents. We turn to this dataset both to show further results of POSLDA's pattern recognition ability along the axes of both semantics and syntax, and to demonstrate the scalability of the model to a larger corpus. Because this is a much larger dataset than the TREC AP corpus, we set the number of topics $K = 50$, but leave the class parameters untouched. Note that this approach to ``guessing'' the number of topics represented by a dataset is a typical way to begin understanding the make-up of a collection of documents. With the parametric version of POSLDA, we can use perplexity as calculated on a held-out test set to help determine the best number of topics and this is explored in the following subsection. For a more principled approach, we can use an HDP prior over the number of topics \cite{hdp}.

\begin{center}
\begin{table}
\begin{center}
\begin{footnotesize}
\begin{tabular}{|c|c|c|c|c|c|}
\hline
\multicolumn{3}{|c}{\textbf{\textit{``energy''}}} & \multicolumn{3}{c|}{\textbf{\textit{``corporations''}}} \\
\textit{adj} & \textit{verb} & \textit{noun} & \textit{adj} & \textit{verb} & \textit{noun} \\
\hline
chemical & clean & plant & executive & named & president\\
power & waste & plants & vice & holding & company\\
environmental & attack & agency & operating & banking & officer\\
water & adore & utility & financial & managing & chairman\\
electric & exist & facility & management & succeed & board\\
energy & create & commission & company & succeeds & directory\\
air & combust & co. & board & named & post\\
pont & protect & industry & former & reacquired & unit\\
safety & mine & environment & investment & created & executive\\
gas & insult & corp. & division & centers & executives\\
\hline
\end{tabular}
\end{footnotesize}
\caption{\label{wsj_examples} Example topics learned from the ACL\_DCI WSJ dataset with POSLDA.}
\end{center}
\end{table}
\end{center}

Table \ref{wsj_examples} shows two topics learned on the WSJ dataset with the parameters described above. Again, we show each topic as three POS-specific topics learned from POSLDA: verb, noun, and adjective. We show some of the more interpretable part-of-speech-specific topics, but in general the learned distributions appear noisier than those found on the smaller AP dataset. Nevertheless, the found topics are subjectively interpretable.

\subsubsection{Domain Specific Data}

While corpora of newswire documents are prevalent in NLP research due to their contained articles' proclivity, there are a number of other domains with large collections of data that will benefit from new statistical models for corpus exploration, data mining, and other text-related tasks. These areas may include, \textit{inter alia}, public health, economics, and law. Studying domain-specific corpora can be illuminating in text modeling research because often domains are filled with eccentricities that do not show up in general writing. These include domain-specific vocabularies and specialized writing structures. Law is a particularly relevant field because of the abundance of textual data that is produced in the field. Here, we demonstrate the qualitative effectiveness of modeling a collection of Supreme Court of Canada decisions with the POSLDA model.

\begin{table}
\begin{center}
\begin{footnotesize}
\begin{tabular}{|c|c|c|c|c|c|}
\hline
\multicolumn{3}{|c|}{\textbf{\textit{``criminal''}}} & \multicolumn{3}{c|}{\textbf{\textit{``labour''}}}\\
\textit{adj} & \textit{verb} & \textit{noun} & \textit{adj} & \textit{verb} & \textit{noun}\\
\hline
criminal & appeal & accused & labour & appeal & employer\\
mens & respect & offence & collective & finding & employee\\
bodily & committing & person & employment & issue & union\\
actus & section & code & union & respect & board\\
reasonable & determining & crown & bargaining & section & code\\
subjective & relating & act & trade & determining & agreement\\
mental & doing & conviction & individual & join & employees\\
objective & carrying & law & employee & colleague & court\\
unlawful & proof & defence & agricultural & lester & relationship\\
common & aiding & crime & construction & dismissing & position\\
\multicolumn{6}{|c|}{}\\
\multicolumn{3}{|c|}{\textbf{\textit{``insurance''}}} & \multicolumn{3}{c|}{\textbf{\textit{``family''}}}\\
\textit{adj} & \textit{verb} & \textit{noun} & \textit{adj} & \textit{verb} & \textit{noun}\\
\hline
insurance & appeal & policy & spousal & appeal & court \\
insured & respect & insurer & child & account & spouse \\
hypothecary & effect & insured & family & determining & parties \\
disability & insure & contract & support & respect & agreement \\
unemployment & defend & clause & economic & regard & marriage \\
exclusion & issue & claim & financial & consider & act \\
life & accord & risk & matrimonial & considering & wife \\
automobile & indemnify & insurance & pension & accord & pension \\
third & force & premium & married & subsection & relationship \\
standard & insuring & beneficiary & marriage & section & husband \\
\hline
\end{tabular}
\end{footnotesize}
\caption{\label{scc_examples} Example topics learned from the Supreme Court of Canada decisions 1989 to 2009 with POSLDA.}
\end{center}
\end{table}

We choose $K=40$, $S = 17$, and $S_{\textsc{SEM}} = 7$, as above. Four of the uncovered topics broken into adjective, verb, and noun are displayed in Table \ref{scc_examples}. Once again, the POS-specific topics are clear and interpretable. In the subtopic for ``criminal law'' adjectives, for example, there are a number of first-words from some common criminal law phrases. These include ``mens'' from the common phrase \emph{mens rea} (the mental component required to commit a crime), ``actus'' from the common phrase \emph{actus reus} (the physical act component required to commit a crime), and ``reasonable'' which often modifies the word ``doubt'' to form the phrase \emph{reasonable doubt}. Furthermore, the nouns in the criminal law topic help make clear that domain specific data have been understood properly. The most probable word in the subtopic for ``criminal law'' nouns is ``accused'' which might na\"{\i}vely be tagged as a verb in a dictionary-only based method that has little understanding of context. Here, it is correctly placed in the \emph{noun} subtopic because the \emph{accused} is the person that stands accused of a crime.\footnote{See, \textit{e.g.} Black's Law Dictionary (2d Pocket ed. 2001).}

As with other datasets there is some noisiness in the results such as the word ``section'' appearing in the verb subtopic for the criminal, labour, and family law topics. This is likely due to the fact that in legal decisions phrases describing the origin of laws typically have their own sort of quasi-grammar and ``section'' is a common word in this context. An example is found in \textit{Winters v. Legal Services Society}\footnote{\textit{Winters v. Legal Services Society}, [1999] 3 S.C.R. 160.} where our sentence splitter found the sentence ``The Requirements of Section 3(2)''. For this to be a proper sentence it requires a verb. The model has likely decided that the word ``section'' would work the best as a verb in this context and it is therefore likely an error attributable to the sentence splitting algorithm as opposed to the POSLDA model. Another problem is that the verb ``appeal'' has shown up as the most important verb in every topic. This is an interesting issue because typically we have seen indistinguishing words pushed to the syntax-only classes. One likely reason that this has not happened is that, though the word is important in many topics, it is not important in \emph{all} of the uncovered latent topics. Appeal is also an interesting word in this domain because it is used to a great extent as both a verb (to \textit{appeal} a decision) and as a noun (the \textit{appeal} in question). While appeal is a common noun, it does \textit{not} show up as any of the top ten nouns in any of the displayed noun subtopics. Despite these slight inconsistencies, however, the POSLDA-learned topics from the SCC dataset are interpretable and clear.

\subsubsection{Noisy Data}

Recently, there has been a large interest in mining the vast quantity of text data created every day though the popular microblogging service Twitter.\footnote{\url{http://www.twitter.com}.} There are a number of unique challenges associated with analyzing this kind of data, however, that make it different from the datasets studied above. First, unlike the highly-structured news articles in corpora such as TREC AP and WSJ, Twitter messages (``tweets'') are rarely well-formed sentences. Proper grammar (or even anything close to it) is all but abandoned on Twitter and the rules of punctuation have been seemingly reinvented \cite{TMtwitter}. One of the principal reasons for this style of writing is that tweets are constrained to a maximum of 140 characters. This of course poses its own issues with respect to text modelling as short documents will contain less thematic information. In addition, partially due to character-limit constraints and partially because Twitter is very popular amongst young people, proper spelling is rare in such a dataset \cite{ritterTwitter}. Each of these issues poses unique problems for modelling topics in this area. Because the long-range thematic dependencies in POSLDA are determined by word co-occurrence, multiple spellings of the same word can hinder unsupervised topic recognition. On the other hand, because the short-range syntactic dependencies in POSLDA are learned by understanding common word-class transitions, structureless grammar also causes problems in determining parts-of-speech.

Despite the issues outlined above, however, Twitter is a very interesting resource. It represents the up-to-the-minute thoughts of millions of people across the world and the knowledge that can be learned from this data is likely immense. One of the most interesting analyses of Twitter data so far is to use a supervised topic model to learn current issues about public health such as what health problems are being experienced by whom (and where) and how people are treating those problems \cite{twitterHealth}. Here, we simply demonstrate that without any additional machinery, POSLDA can learn semantically and syntactically consistent topics from a collection of tweets. We use the Twitter POS dataset released at ACL 2011 by Gimpel, et al.\ which consists of approximately 26,000 words across 1,827 tweets \cite{twitterpos}. While this is a fairly limited collection of data, our model is nevertheless able to uncover some interesting POS-specific topics. Table \ref{twitter_examples} shows three manually-labeled topics learned with the settings of $K = 20$, $S = 17$, and $S_{\textsc{SEM}} = 7$.

\begin{table}
\begin{center}
\begin{footnotesize}
\begin{tabular}{|c|c|c|c|c|c|}
\hline
\multicolumn{2}{|c|}{\textbf{\textit{``media''}}} & \multicolumn{2}{c|}{\textbf{\textit{``relationships'' (?)}}} & \multicolumn{2}{c|}{\textbf{\textit{``movies'' (?)}}} \\
\textit{verb} & \textit{noun} & \textit{verb} & \textit{noun} & \textit{verb} & \textit{noun} \\
\hline
phone & photo & spent & girl & watch & online\\
posted & video & headed & party & looking & movie\\
speak & smile & cheating & sex & bored & script\\
looking & phone & visit & bitches & seen & series\\
send & night & annoy & sleep & walk & avatar\\
follow & time & callin & eyes & thats & funny\\
listen & media & wanna & love & respect & cool\\
chat & happy & hang & rings & cleared & bad\\
love & chick & f*ck & games & announce & girl\\ 
heart & da & sucks & date & wow & dollar\\
\hline
\end{tabular}
\end{footnotesize}
\caption{\label{twitter_examples} Example topics learned from Twitter with POSLDA.}
\end{center}
\end{table}

The topics outlined in Table \ref{twitter_examples} are by far the noisiest and least interpretable demonstrated thus far. Because of the non-standard text structure and issues with incorrectly spelled words, it is difficult for the algorithm to uncover the patterns of interest. Nevertheless, of the 20 topics, three fairly interpretable topics are demonstrated. As it is difficult to tell exactly which part-of-speech each subtopic represents, we only list two distinct subtopics per topic: verb and noun. The first topic -- ``media'' -- is likely the most coherent while the other two have been labelled with trailing question marks to convey that it is difficult to name the topics. It is for this reason that Twitter-specific topic models (or at least Twitter-specific alterations to existing topic models) may be required \cite{TMtwitter,ritterTwitter}. Analyzing the listings in Table \ref{twitter_examples} does show, however, that POSLDA can uncover interesting semantic and syntactic patterns even in this highly noisy source.

\subsection{Quantitative Results}

There are several methods that are commonly employed to evaluate novel probabilistic models in the literature \cite{evaltm}. The original LDA paper -- and many others -- use the \emph{perplexity} which is a standard metric in the information retrieval literature \cite{lda,hdp}. A probabilistic model can also be evaluated by considering its performance on an extrinsic task. Here, we first focus on the perplexity and use it to measure POSLDA's performance as a predictive language model, and then discuss using the model for unsupervised POS tagging.

\subsubsection{Predictive Language Modelling}

Following the standard practice in topic modeling research \cite{lda,hmmlda,hdp,taglda}, we fit a model to a training set and compute the perplexity of a held-out test set. The perplexity can be defined as the predicted average number of words that are equally likely to be generated for a given position \cite{taglda}. It is also a monotonically decreasing function of the log likelihood.

The perplexity of a held-out test set $\mathcal{D}_{test} = \left(\mathbf{w}_d \right)_{d=1}^{M}$ is given as:
\begin{equation}
ppx(\mathcal{D}_{test}) = \exp\left(- \frac{\sum_{d=1}^M{\log p(\mathbf{w}_d | \mathcal{M})}}{\sum_{d=1}^{M}{N_d}}\right)
\end{equation}
where $\mathcal{M}$ represents the model parameters learned from the training data, $p(\mathbf{w}_d | \mathcal{M})$ is the probability (likelihood) of document $\mathbf{w}_d$ given the learned model parameters, and $N_d$ is the number of words in document $\mathbf{w}_d$.

We test the POSLDA model on a subset of the AP TREC dataset. We use ten-fold cross-validation where the data is split into 10 subsets of equal size. We conduct 10 experiments where one of the subsets is held out for testing and the model is trained on the remaining 9 subsets. We report the average over the 10 experiments. We compare the perplexity of POSLDA to the original LDA model, a Bayesian HMM, and Griffiths, et al.'s HMMLDA. Each model is trained using 10,000 iterations of Gibbs sampling. We use asymmetric priors on the document-topic portions $\boldsymbol{\theta}$ and the rows of the HMM transition matrix $\boldsymbol{\pi}$. Following \cite{rethinking}, we use a symmetric prior on the topic-word distributions $\boldsymbol{\phi}$ because having asymmetric values on the topics themselves does not seem to lead to improved performance. The prior's hyperparameters are optimized using Minka's fixed-point method \cite{wallachphd}. For these experiments we use $S=10$ classes of which 5 are designated as semantic. By definition, the HMMLDA model has $SS=1$ and the LDA model has $S=S_{\textsc{sem}}=1$. The HMM model does not consider the number of topics $K$.

\begin{figure}[htbp]
\begin{center}
\includegraphics[width=3in]{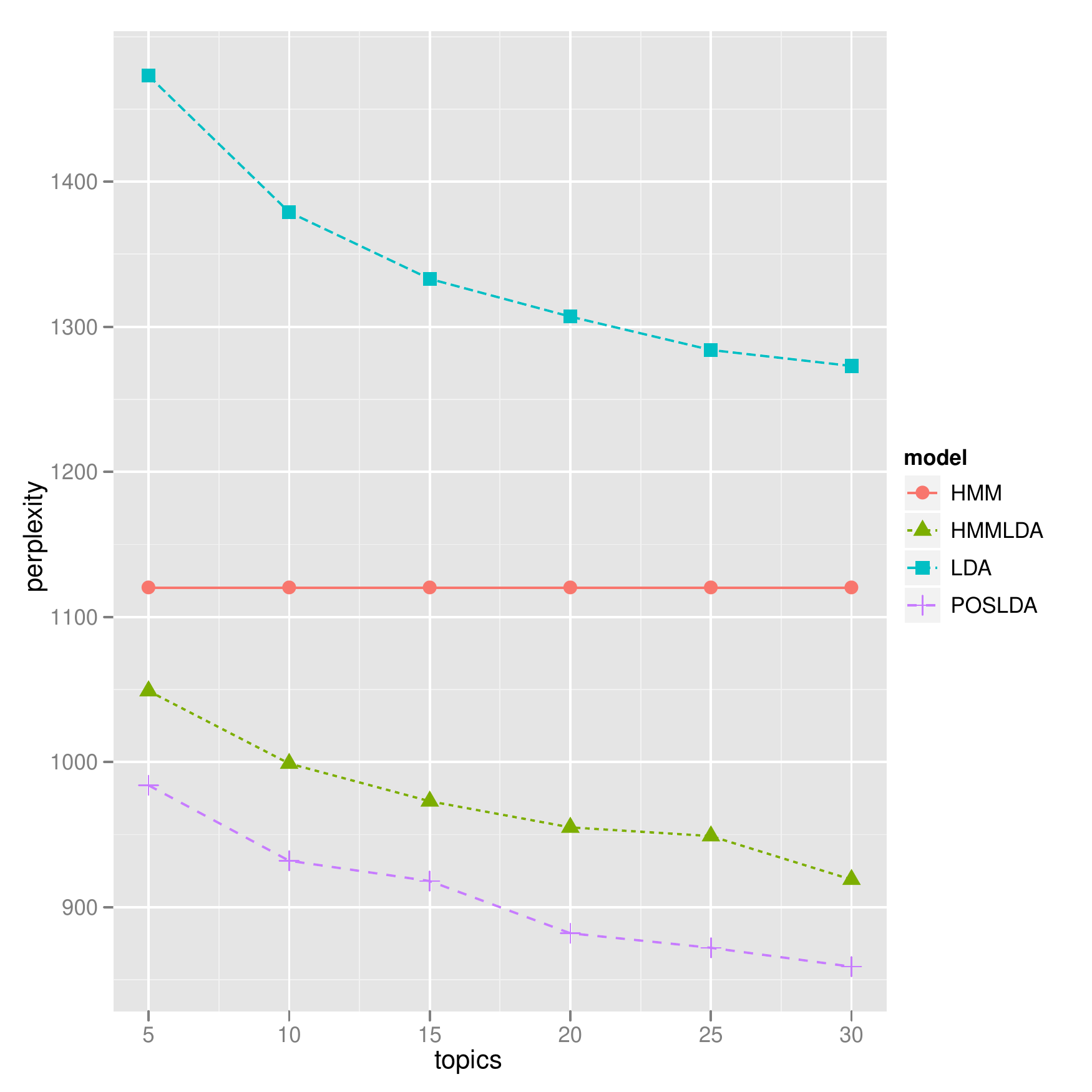}
\end{center}
\caption{\label{fig:ppx_topics} Perplexity of POSLDA and other similar probabilistic models as $K$ varies.}
\end{figure}

Figure \ref{fig:ppx_topics} shows the average perplexity values on a held-out test set for a number of models in the same family as POSLDA over a range of topic values. The HMM achieves lower perplexity values than LDA at all topic settings $K=\{5,10,15,20,25,30\}$. All three topic-based models realize perplexity improvements as the number of topics $K$ increases. Both HMMLDA and POSLDA -- which combine the benefits of the HMM and LDA models -- result in lower perplexity values. POSLDA's additional flexibility in terms of the additional semantic classes allows it to record the lowest perplexity values of all the models tested for each topic setting.

\begin{figure}[htbp]
\begin{center}
\includegraphics[width=3in]{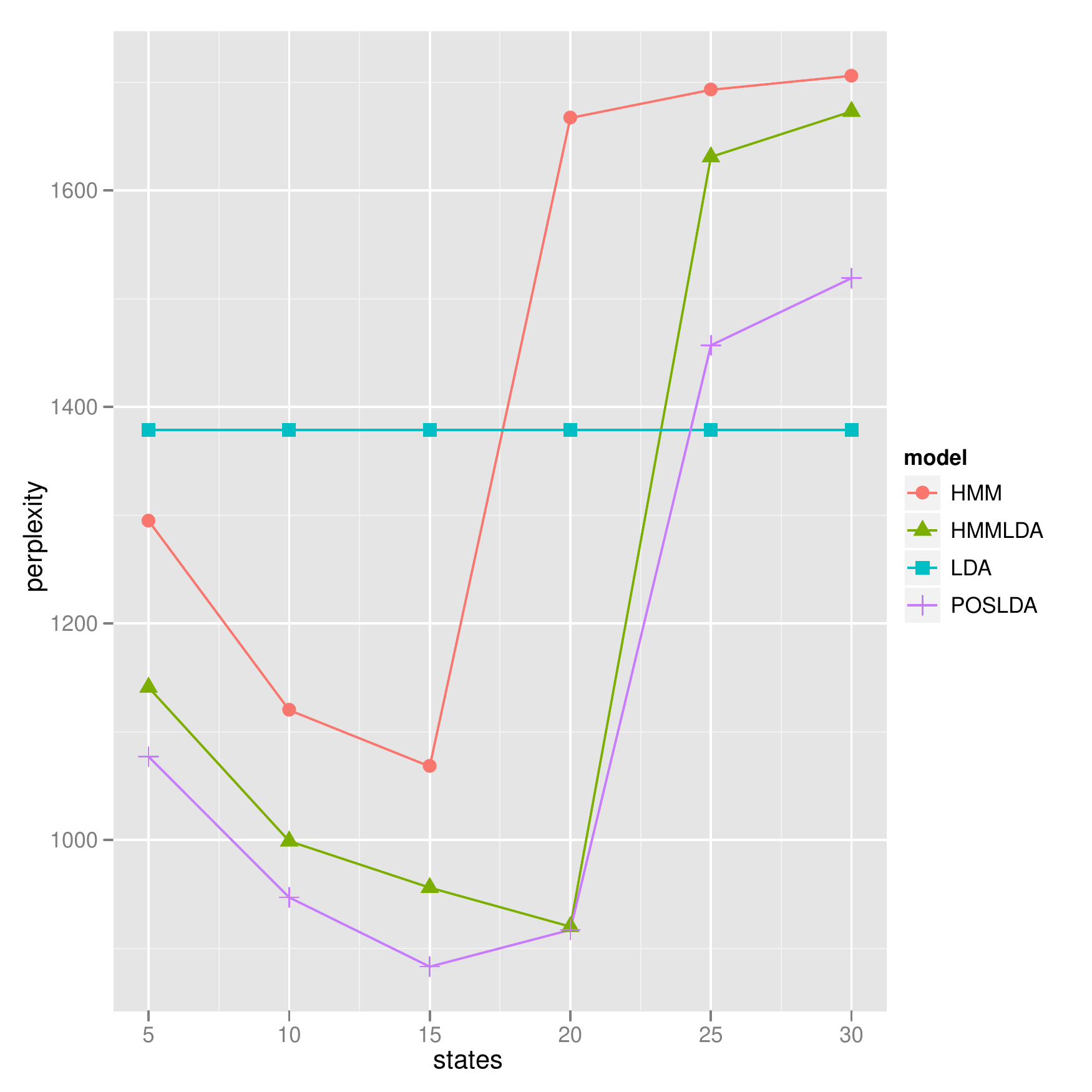}
\end{center}
\caption{\label{fig:ppx_states} Perplexity of POSLDA and other similar probabilistic models as $S$ varies.}
\end{figure}

While Figure \ref{fig:ppx_topics} shows how the POSLDA model's ability to generalize on unseen data is affected by the number of topics, Figure \ref{fig:ppx_states} illustrates the changes in the perplexity when the number of classes $S$ is the independent variable. While it may not reflect the best possible values for the POSLDA model, we set the number of semantic classes $S_{\textsc{sem}}=S$. Interestingly, the HMM's perplexity starts to shoot up when $S=15$, and both HMMLDA and POSLDA show poor perplexity when $S=20$. This likely reflects too much variation in the model (especially with no disambiguation between semantic and syntactic classes). Next, we look at how the perplexity varies when $S$ is held fixed but $S_{\textsc{sem}}$ is free to vary.

\begin{figure}[htbp]
\begin{center}
\includegraphics[width=3in]{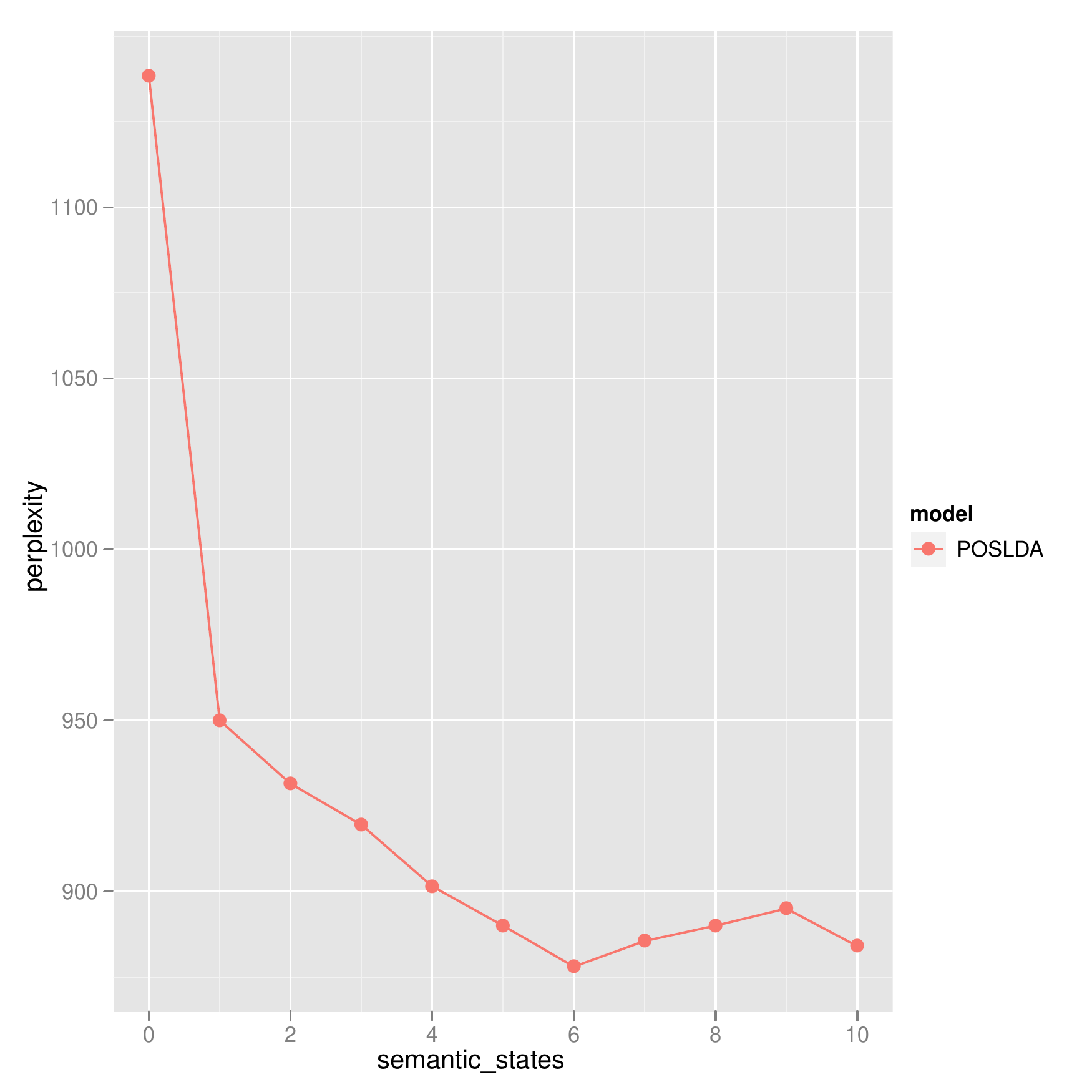}
\end{center}
\caption{\label{fig:ppx_SS} Perplexity of POSLDA as $S_{\textsc{sem}}$ varies.}
\end{figure}

Figure \ref{fig:ppx_SS} shows the average perplexity as the number of semantic states $S_{\textsc{sem}}$ is varied. We set $S=10$ and investigate how different settings of $S_{\textsc{sem}}$ affect the model's ability to generalize. When $S_{\textsc{sem}}=0$ the model reduces to a Bayesian HMM and when $S_{\textsc{sem}}=1$ it becomes the HMMLDA model. Adding some semantic information with HMMLDA improves the perplexity considerably (1172 to 957) and further distinguishing semantic information in POSLDA continues to improve the perplexity until it reaches a minimum at $S_{\textsc{sem}}=6$. This reflects the fact that certain classes of words such as conjunctions are not aided by thematic information.

Above we have demonstrated both that POSLDA tends to learn interpretable topics that are part-of-speech specific, and that it can lead to better predictive performance than other similar generative probabilistic models. The purposes of the models are clearly different, however. POSLDA is more flexible, but it requires a more expensive representation of text (a sequence of words rather than a bag). HMMs are also extremely useful and versatile for a number of applications not necessarily related to text. Next we are interested in applications where the POSLDA model is truly needed or can truly make a difference. In \cite{darling11} we showed how a POSLDA-like model can improve the results in a text summarization task. In the next section, we concentrate on unsupervised part-of-speech tagging.

\subsection{Unsupervised POS Tagging}

Goldwater and Griffiths show that Bayesian HMMs increase the accuracy of unsupervised POS tagging by up to 14 percentage points over the MLE approach \cite{bayesianhmm}. While these results are impressive, unsupervised approaches continue to fall well short of the accuracy obtained with supervised taggers. Nevertheless, unsupervised approaches are preferred in many situations especially when there is no access to large quantities of training data in a specific domain or particular language. We therefore aim to continue improving accuracy with unsupervised approaches by introducing semantics as an additional source of information for this task.

The word ``seal'' appears both as a verb (to \emph{seal} a jar or a leak) and as a noun (the marine mammal \emph{Pinniped}) in the WSJ treebank dataset. The HMM approach can often tag each of these occurrences appropriately given the context, but there are cases where it will fail. However, if the topic being discussed is marine biology, we have another piece of evidence that increases the likelihood that this occurrence of the word ``seal'' is a noun about the marine mammal. If the topic is about pickling or roofing, however, ``seal'' as a verb is given more evidence. Another example is the word ``book''. In a literary context it will almost always take the form of a noun. However, in a topic about promotions or services, the word is more likely to function as a verb: ``to \emph{book} a hotel''. Following this intuition, we use the POSLDA model with the number of HMM states set to the number of possible tags in the tag set and use the state index learned through posterior inference as the predicted tag for each word.

We take two approaches to perform unsupervised POS tagging using the POSLDA model. The first approach uses a tag dictionary containing varying amounts of information on the possible tags that certain words have taken on in the training data. This renders the problem to a case of POS disambiguation rather than pure unsupervised tagging. It is, however, the most common approach to demonstrating results in unsupervised tagging \cite{bayesianhmm}. The second approach is a pure unsupervised method that implements POS clustering. Because we cannot know which classes represent which parts-of-speech, we instead compare the learned clusters to the correct clustering where clusters are exchangeable. Accordingly, we use the variation of information (VI) for evaluating this task \cite{meila}.

We showed in \cite{darling2012} that this model consistently beats the Bayesian HMM approach in the Twitter domain. Here, we show that these improvements hold in a more traditional domain: the ACL\_DCI release of the Penn Treebank's collection of Wall Street Journal newswire articles. This dataset contains approximately 3M words over 6K documents. We condense the tag set to the more standard for unsupervised POS tagging 17-tag set introduced by \cite{smitheisner}.

We follow the established form of \cite{merialdohmm} and \cite{bayesianhmm} for unsupervised POS tagging by making use of a tag dictionary to constrain the possible tag choices for each word and therefore render the problem closer to disambiguation. Like in \cite{bayesianhmm}, we employ a number of dictionaries with varying degrees of knowledge. A dictionary contains the tag information for a word only when it appears more than $d$ times in the training corpus. We ran experiments for $d = 1, 2, 3, 5, 10,$ and $\infty$ where the problem becomes POS clustering. We report both tagging accuracy and the variation of information (VI), which computes the information lost in moving from one clustering $C$ to another $C'$: $VI(C,C') = H(C) + H(C') - 2I(C,C')$ \cite{meila}. This can be interpreted as a measure of similarity between the clusterings, where a smaller value indicates higher similarity.

To properly make use of a tag dictionary, we slightly alter the POSLDA model by adding an additional observed random variable to the model as in Labeled LDA \cite{labeledlda}. The tag dictionary is encoded in the model by a list of binary class indicators for each word $\Lambda^{(w)} = (\lambda_{w_1}, ... , \lambda_{w_V})$ where $\lambda_w = (l_1, ..., l_C)$ for each class. The model then becomes as depicted in Figure \ref{fig:poslda_gm} (right). The sampling equation is then updated simply to $p(z_i, c_i | \bm{c_{-i}}, \bm{z_{-i}}, \bm{w}) \propto p(z_i, c_i | \bm{c_{-i}}, \bm{z_{-i}}, \bm{w}) \times \lambda_{w, c_i}$.

We run our Gibbs sampler for 20,000 iterations and obtain a maximum a posteriori (MAP) estimate for each word's tag by employing simulated annealing. Each posterior probability $p(c, z | \cdot)$ in the sampling distribution is raised to the power of $\frac{1}{\tau}$ where $\tau$ is a value (in traditional annealing used in physics $\tau$ would be a temperature) that approaches $0$ as the sampler converges. This approach is akin to bringing a system from an arbitrary state to one with the lowest energy thus viewing the Gibbs sampling procedure as a random search whose goal is to identify the MAP tag sequence, as employed in \cite{bayesianhmm}. Finally, we run each experiment 5 times from random initializations and report the average accuracy and variation of information. All improved results reported for the POSLDA model are statistically significantly different from those achieved with the BHMM model as determined by a Student's t-test with $p \ll 0.01$.

The achieved results are shown in Table \ref{WSJResults} for a random tag assignment, the Bayesian HMM (BHMM) described in \cite{bayesianhmm}, and our POSLDA tagging approach. For both BHMM and POSLDA, we optimize the asymmetric hyperparameters $\alpha$ and $\gamma$ and the symmetric prior $\beta$ using either direct optimization or sampling with similar results. We use 6 semantic classes -- adjective, verb, gerund or present participle verb, adverb, noun, and past participle verb -- leaving 11 remaining for syntax words, and we report results for $K = 10$ topics.

\begin{table}
\begin{center}
\begin{tabular}{c|c|c|c|c|c|c}
Accuracy & 1 & 2 & 3 & 5 & 10 & $\infty$\\
\hline
random & 58.7 & 57.9 & 57.5 & 56.7 & 55.5 & \\
BHMM & 86.4 & 85.6 & 85.3 & 84.9 & 84.5 & \\
POSLDA & \textbf{87.7} & \textbf{87.2} & \textbf{86.9} & \textbf{86.5} & \textbf{85.9} & \\
\hline
VI & & & & & &\\
\hline
random & 2.37 & 2.44 & 2.48 & 2.54 & 2.63 & 5.07\\
BHMM & 0.77 & 0.83 & 0.86 & 0.90 & 0.90 & 2.25\\
POSLDA & \textbf{0.73} & \textbf{0.77} & \textbf{0.79} & \textbf{0.82} & \textbf{0.86} & 2.30\\
\hline
Corpus stats & & & & & &\\
\hline
\% ambig. & 66.0 & 66.8 & 67.4 & 68.1 & 69.3 & 100\\
tags / token & 2.34 & 2.48 & 2.57 & 2.71 & 2.95 & 17\\
\end{tabular}
\end{center}
\caption{\label{WSJResults} POS Tagging Results on WSJ dataset.}
\end{table}

\begin{figure}[htbp]
\begin{center}
\includegraphics[width=3in]{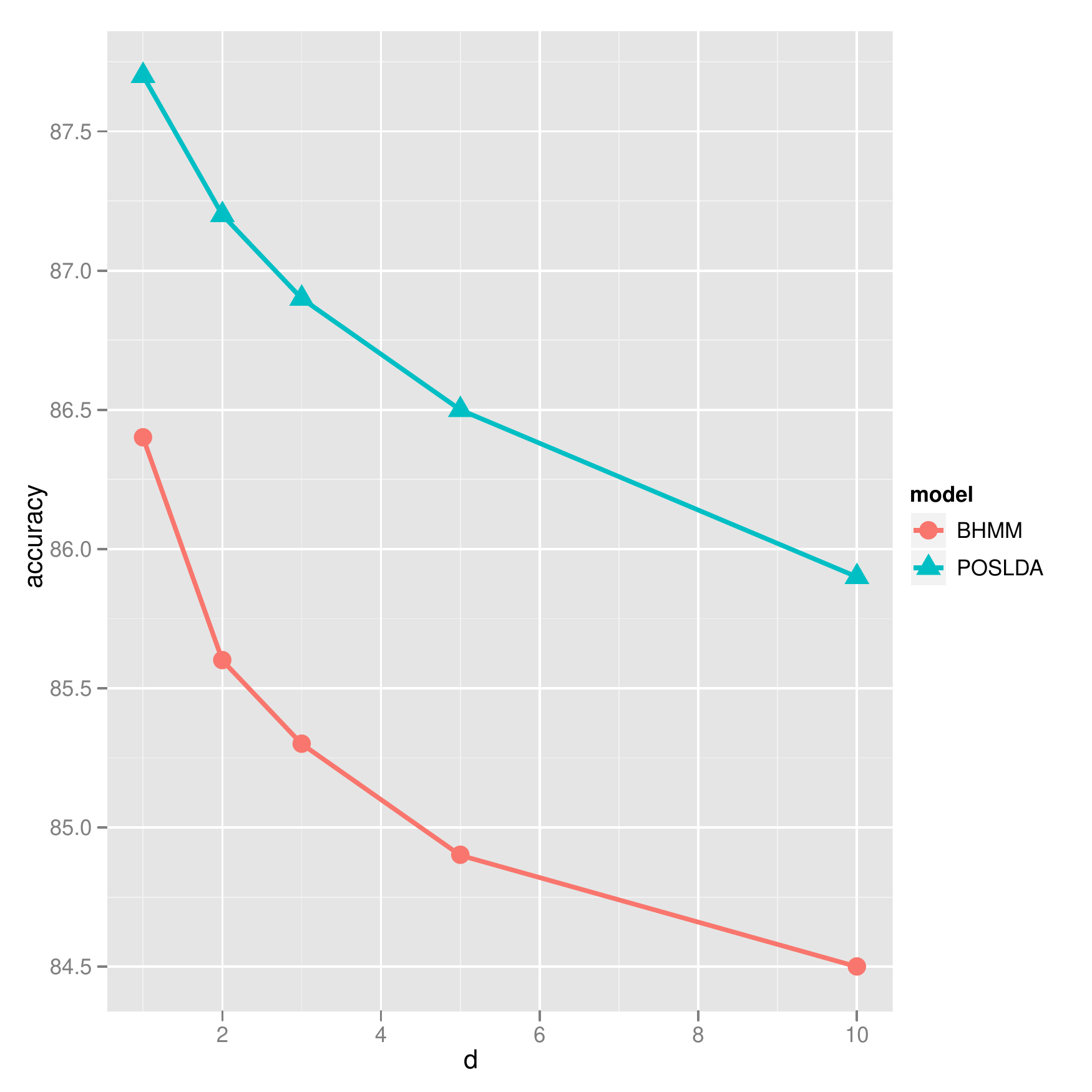}
\end{center}
\caption{\label{WSJaccuracy} POS Tagging Accuracy on WSJ dataset.}
\end{figure}

These POS tagging results are also shown graphically in Figure \ref{WSJaccuracy}. Our model outperforms the Bayesian HMM for all values of $d$ in accuracy and all values of $d$ but one for variation of information. In fact, POSLDA maintains higher tagging accuracy for $d = 5$ than the Bayesian HMM for $d = 1$. Our approach consistently surpasses the results achieved with BHMM by approximately 1.5 percentage points for each value of $d$. The one case where POSLDA did not improve upon BHMM is where $d = \infty$ for POS clustering.

\section{Conclusions and Future Work}

In this article we presented the combined topic and syntax model, \textit{Part-of-Speech} LDA or POSLDA. We have also demonstrated its use as an improved model for performing unsupervised POS tagging. Our overarching goal is to demonstrate that combining the two axes of word meaning -- syntax and semantics -- into a coherent model can result in improvements to both learned topic distributions and to NLP tasks such as POS tagging. We showed that incorporating semantic information into the HMM model led to improved results for this task. Additionally, we showed that combining the two axes of word information results in a language model that achieves lower perplexity -- and therefore better predictive capability -- than other similar probabilistic models.

In future work we would like to apply the POSLDA model to other NLP tasks that also rely upon learned word distributions. These include text summarization, text segmentation, and translation. An interesting avenue for further research is POSLDA's ability to serve as the base of a language generation system. While performing the LDA generative process would perhaps result in documents that contain words that are semantically related, we cannot say that it is generating language. POSLDA on the other hand -- with some strong prior knowledge -- may be able to generate coherent natural language because it exhibits more structure in the language generation process. We hope to explore this direction of research and apply it to tasks such as database population and abstract text summarization.

\begin{acknowledgments}
The authors would like to thank the Natural Sciences and Engineering Research Council of Canada for partially funding this work through a Doctoral Post-Graduate Scholarship for William Darling. The authors would also like to acknowledge the financial support provided by Ontario Centres of Excellence (OCE) through the OCE/Precarn Alliance Program.
\end{acknowledgments}

\bibliographystyle{fullname}
\bibliography{newposlda}

\end{document}